\documentclass[
  journal=chr,
  manuscript=Registered\_Report\_Protocol,
  year=2025,
  volume=1,
]{cup-journal}

\usepackage{amsmath}
\usepackage[nopatch]{microtype}
\usepackage{booktabs}

\usepackage{tabto}
\usepackage[hidelinks]{hyperref}

\title{It takes a village to write a book: \\mapping anonymous contributions in Stephen Langton's Quaestiones Theologiae}

\author{Jan Maliszewski}
\affiliation{Faculty of Philosophy, University of Warsaw\\ j.maliszewski@uw.edu.pl}

\keywords{stylometry, handwritten text recognition, reportationes, scholasticism, Stephen Langton} 

\begin{document}

\begin{abstract}

While the indirect evidence suggests that already in the early scholastic period the literary production based on records of oral teaching (so-called \textit{reportationes}) was not uncommon, there are very few sources commenting on the practice. This paper details the design of a study applying stylometric techniques of authorship attribution to a collection developed from \textit{reportationes} --- Stephen Langton's \textit{Quaestiones Theologiae} --- aiming to uncover layers of editorial work and thus validate some hypotheses regarding the collection's formation. Following \citet{CampsCléricePinche2021}, I discuss the implementation of an HTR pipeline and stylometric analysis based on the most frequent words, POS tags, and pseudo-affixes. The proposed study will offer two methodological gains relevant to computational research on the scholastic tradition: it will directly compare performance on manually composed and automatically extracted data, and it will test the validity of transformer-based OCR and automated transcription alignment for workflows applied to scholastic Latin corpora. If successful, this study will provide an easily reusable template for the exploratory analysis of collaborative literary production stemming from medieval universities.

\end{abstract}



\section*{Plain Language Summary}

Many texts produced at the medieval universities did not originate as literary works but were instead gradually and collaboratively developed from records of oral teaching, known as \textit{reportationes}. While this practice was likely widespread, there are very few sources detailing its daily operation, forcing scholars to rely on indirect evidence deducible from preserved works. In this context, this paper proposes a study exploring computational analysis of style as a way to track layers of editorial work in scholastic collections, potentially revealing the actual scope of authors' control over these works. This approach draws from earlier studies which successfully employed computational techniques in the context of medieval Latin letter collections and Old French hagiographies. I discuss applying similar methods to the collection of Stephen Langton's (d.~1228) theological \textit{quaestiones}. Langton's collection is particularly interesting for it is known to depend on \textit{reportationes}, and it transmits most of its material in more than one version, in some cases allowing us to track the development from raw records of oral teaching to fully developed literary forms. Initial analysis of Langton's corpus shows that by measuring the frequencies of the most common words --- a common stylometric method --- it is possible to differentiate its stylistic signal from other contemporary scholastic collections, as well as to observe some stylistic diversity within Langton's corpus. However, the key limitation in the context of Langton's \textit{quaestiones} stems from their length, as most of \textit{quaestiones} are too short to provide representative samples. This issue can be addressed by including additional stylistic features: sequences of Part of Speech tags, which capture syntactic structures, and pseudo-affixes (the few opening and closing characters of each word), which represent morphological information. These features have been shown to provide good results with automatically generated transcriptions; consequently, I plan to compare tests performed on manually composed editions and automatically extracted data. The key gain offered by automated transcription lies in providing a feasible way of extending analysed corpora by including unedited material.

\CUPTWOCOL

\section{Introduction}

This paper proposes a study employing stylometric techniques of authorship attribution to assess the scope of anonymous contributions to the collection of Stephen Langton's Quaestiones Theologiae. In this, it follows studies which demonstrated the robustness of stylometric methods applied to the analysis of collaborative authorship in comparable medieval Latin literary traditions (\cite{KestemontMoensDeploige2013}; \cite{DeGussem2017}). In particular, I draw heavily on the methods of unsupervised cluster analysis offered in \citet{CampsCafiero2013}, \citet{CafieroCamps2019}, \citet{CampsCléricePinche2021}. The central goal of the proposed study is to analyse stylistic signals observable within a collection known to originate from anonymous \textit{reportationes} --- the collection of Stephen Langton’s \textit{Quaestiones Theologiae} --- aiming to locate any internal stylistic clusters. The hypothesis is that, if discernible, such clusters may be representative of the activity of non-authorial contributors. While the proposed study's design is informed by recent editorial work on Langton's collection (Langton, ed.~Bieniak et al.~2014--2024), these methods can be expected to apply to other scholastic corpora displaying similar traces of collaborative work. To further explore this potential transfer of methods, the proposed study will involve a direct comparison of the performance of the stylometric tests on both manually edited and HTR-extracted data, adapting the pipeline constructed in \citet{CampsCléricePinche2021}. Below, I discuss the philological motivation of the problem, followed by a discussion of the selected methods and potential results.

\subsection{State of research on early scholastic reportationes}

Dating back at least to the 1920s, the scholarly interest in the production of \textit{reportationes} gradually led to their recognition as a salient feature of the scholastic intellectual practice.\footnote{For a historical summary of the literature on \textit{reportationes}, see \cite[p.~74--76]{Saccenti_Riccardo2016}.} Generally speaking, a \textit{reportatio} is a note recording oral teaching, usually taken from a master’s lecture by one of its participants. The proliferation of \textit{reportationes} was closely associated with the growth of universities, and many attempts were made to analyse \textit{reportationes} in the context of specifically medieval didactic forms. Thus, for example, \textit{reportationes} prove uniquely valuable as testimonies of the practice of formal public debate, \textit{disputatio}, in the 13th and 14th centuries.\footnote{See \cite[p.~420]{Hamesse1997}. For a comprehensive study of the practice of \textit{disputatio}, see \cite{Weijers2013}.} Still, \textit{reportatio} as such was neither a genre nor a transmission method but a technique applied in many different contexts and with varying aims.\footnote{As commonly acknowledged after \cite{Hamesse1997}. A notable context outside of this study's scope is sermon \textit{reportationes} --- see \citet{Roberts1968}, \citet{dAvray1985}, \citet{Beriou}.} In many cases, the primary goal of such note-taking may have been private, intended to aid the student's memory. However, there are also documented cases in which the teaching collected through \textit{reportationes} formed the foundation of a master’s regular literary works. It is not always easy to establish whether a particular text originated from \textit{reportationes}, and thus the scope of such oral-to-literary transfer is not fully understood. While the literary production based on \textit{reportationes} dates back at least to the 1120s, for the entire 12th century scholars have identified only two testimonies describing the process of reporting and its later literary refinement.\footnote{These testimonies come from Abelard's account of his exegetical lectures (Abelard, ed. Monfrin 1959, pp.~69--70), and from a letter of an otherwise unknown Laurentius, the \textit{reportator} of Hugh of Saint-Victor's \textit{Sententiae de divinitate} (Hugh of St. Victor, ed. Piazzoni, 1982, pp.~912--3). For discussion of these passages, see \citet{Siri2013}, \citet[pp.~16--29]{Foley2024}.\label{footnote_label_1}} Consequently, the existing research on the earliest usage of \textit{reportationes} for literary production --- that is, the production stemming from the cathedral schools and universities before c.~1250 --- largely extrapolates from these two testimonies and the more comprehensive information available for later scholastic tradition. 

Two basic types of evidence provide insight into the actual scope of the early scholastic literary production based on \textit{reportationes}. First, scholars identified marks of oral communication in some otherwise inconspicuous literary works. These marks can be lexical or pragmatic. Examples include the prevalence of second-person verb forms, ellipses, or context-specific references to the audience --- e.g. singling out lecture participants by name or recalling earlier exchanges of arguments, not preserved in the written testimony.\footnote{For a comprehensive discussion of markers of orality preserved in 12th-century collections, see \cite{Siri2013}.} Another type of indirect evidence is stemmatical. It is not uncommon for traditions dating back to 12th-century Paris to transmit multiple partially collatable versions, likely indicating independent strands of transmission in the text's early history. Transmission via \textit{reportationes} is a likely cause behind at least some of this variance,\footnote{Other likely factors shaping irregular transmission in this period include evolution of the text after its initial circulation --- both authorized by the master and independent, e.g. by incorporation of external glosses --- and transmission \textit{per pecia}, i.e. the practice of copying long works from smaller booklets, which may have easily resulted in the circulation of incomplete witnesses. On \textit{reportationes}, dictation, and the practice of transmission \textit{per pecia} in medieval Paris, see \cite[p.~165--174]{Weijers2015}.} especially when more than one record of a lecture was created and when the master did not supervise the process. Taken together, available evidence suggests that already in the early stages of the scholastic tradition, it was fairly common for a master to produce his works from \textit{reportationes}.

Different general accounts of the practice of \textit{reportatio} can be largely traced back to scholars’ interest in corpora exhibiting different consequences of transmission via \textit{reportationes}. Some collections, while demonstrably stemming from classroom reports, are stemmatically regular --- that is, the stemmatical evidence suggests the existence of a single archetype at the origin of the tradition --- leading their editors to assume a higher degree of reportatorial professionalization and master’s control over the process.\footnote{An example of such a regular 12th-century collection developed from \textit{reportationes} can be found in Peter Comestor's Gospel glosses --- see Peter Comestor, ed Foley, 2024, especially the introductory discussion on pp.~17--20.} On the other end of the spectrum, we find collections compiling and reworking scattered reportatorial material, possibly with little or no magisterial control, and at a considerable time distance from the initial lecture.\footnote{This, as discussed below, is the case of Stephen Langton’s \textit{Quaestiones}.} Overall, the \textit{reportatio} seems to be less of a formalized and unified phenomenon in the 12th century than in its later practice, and thus many basic questions relating to its operation remain open. In particular, in most cases we do not know how many actors --- and with what exact roles --- stand behind the preserved collections. A model transmission would involve the reportator reworking his record shortly after the class or debate, presumably mostly to supplement the details missing due to the hastiness of the initial record,\footnote{It should be noted that any preserved record is virtually never identical with the initial \textit{reportatio} since, as far as we know, these were ordinarily produced on provisional writing support, e.g. wax tablets or loose offcuts of parchment. Moving such text to regular parchment folios likely involved at least a minimal degree of editorial normalization.} and then the master authenticating the testimony, likely extensively interfering in the text --- this final correction is known as an \textit{ordinatio}.\footnote{This is the process described by Laurentius, Hugh of St. Victor's pupil reporting \textit{Sententiae de divinitate} (see the reference in n.~\ref{footnote_label_1} above).} How closely the daily operation of textual production based on \textit{reportationes} resembled this schema is not clear, but we can safely assume that the preserved records are skewed on the side of more regular instances of reporting, as these were more likely to enter into wider circulation requiring ample scribal work. 

\subsection{Corpus: Stephen Langton’s Quaestiones Theologiae}
The collection of Stephen Langton’s Quaestiones provides a particularly convenient vantage point for the study of the practice of \textit{reportatio} in the early university setting. Stemming from Langton’s Parisian teaching sometime during the last decades of the 12th century up to 1206, this collection was never given a final shape, despite some clear traces of attempted editorial work. Around 70\% of the \textit{quaestiones} listed in the contemporary index of the collection are transmitted in multiple substantially different versions, preserved at varying stages of production.\footnote{Of the 173 \textit{quaestiones}, 119 are transmitted in two to five different versions. These numbers do not account for the so-called \textit{quaestiones extra indicem}; including these texts and all the versions, the collection contains over 350 different texts. For the complex issues of cataloguing Langtons' \textit{quaestiones}, see \citet{quinto1994doctor} and the introduction to the first volume of the critical edition of Langton's collection, ed. Quinto, Bieniak (2014).} Some of these include exceptionally concise discussions – presumably unedited transcripts of \textit{reportationes} – which correspond with some of the fully developed \textit{quaestiones}, either preserving the structure of the argumentation or being partially collatable, suggesting that these versions represent different accounts of one oral \textit{quaestio}.

The collection is transmitted by eight major manuscript witnesses (Figure 1). The discernible subcollections (mss. C,\footnote{Ms. C consists of six distinct codicological units, Ca--Cf, which occupy different positions in the stemma.} H\,/\,K, and families {$\alpha$} and {$\beta$}) likely represent parallel, partially overlapping compilations of Langton's material. They transmit vastly different sets of \textit{quaestiones}, mostly in varying order. Part of the collection may have been reviewed by Langton --- especially in ms.~C ---  but most of the \textit{quaestiones} were almost certainly edited by someone else, possibly by unknown students or secretaries from Langton's milieu after 1206. How many editors worked on this collection remains unclear. Similarly, we have no estimate of the number of \textit{reportatores} involved in recording Langton's teaching. 

\subsection{Exploratory stylometric analysis}

The basic premise of the proposed study stems from the results of \citet{KestemontMoensDeploige2013} and \citet{DeGussem2017}. Both these studies applied techniques of stylometric authorship attribution in the context of 12th-century collaborative Latin writing, showing that it is possible to track with these tools stylistic variance which can be linked to the contributions of secretaries working with, respectively, Hildegard of Bingen and Bernard of Clairvaux. Our hypothesis --- to some extent validated by the exploratory analysis --- is that it is similarly possible to map the layers of reportatorial and editorial activity in scholastic corpora.

\begin{figure}[hbt!]
\centering
\includegraphics[width=0.75\linewidth]{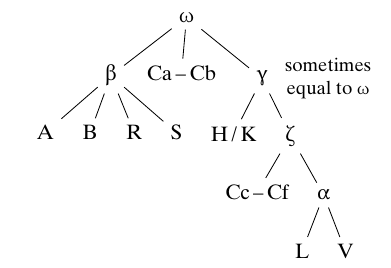}
\caption{General transmission pattern of Langton's Quaestiones Theologiae.}
\label{fig_1}
\end{figure}

Both these studies employed to a good effect a widely accepted metric of style: the frequencies of function words, that is, the most common subject-independent lemmas observed in a given corpus.\footnote{For example, the ten most frequent words (unlemmatized) in Langton's corpus are 'est', 'et', 'non', 'quod', 'in', 'ergo', 'set', 'ad', 'quia', and 'hoc'.} While, as discussed below, the specific stylometric tests applied in these studies do not transfer well into the problem at hand, it can certainly be confirmed that function words provide a reliable marker of style for scholastic corpora. For example, figure 2 shows a comparison of 3,000-word samples from Langton's \textit{quaestiones}, Robert of Courson's Summa,\footnote{On Robert's Summa, see \citet{Kennedy1947}. I used a transcription of ms.~Bruges 247, ff. 4\textsuperscript{va}--61\textsuperscript{va}, kindly shared by Gary Macy.} and Aquinas' Summa Theologiae, \textit{prima pars}.\footnote{Summa Theologiae, I\textsuperscript{a}, qq. 1--45, following the text of Corpus Thomisticum.} From each text, we draw 50 continuous samples. All samples are represented by the relative frequencies of the 200 most frequent words (unlemmatised), which largely align with function words. The data was transformed by primary component analysis (PCA), with the two top components capturing a little over 25\% of the total variance. As apparent in the plot, all samples cluster according to their text of origin, showing that these authorial signals can be identified based on the usage of the most frequent words. It is not surprising --- function words prove to be effective across many languages and genres --- but also not entirely trivial, since theological \textit{quaestiones} of the period \pagebreak belong to a highly technical and formulaic genre, and thus can be expected to display overall fainter stylistic signals than the related epistolary or sermon corpora.

\begin{figure}[hbt]
\centering
\includegraphics[width=1\linewidth]{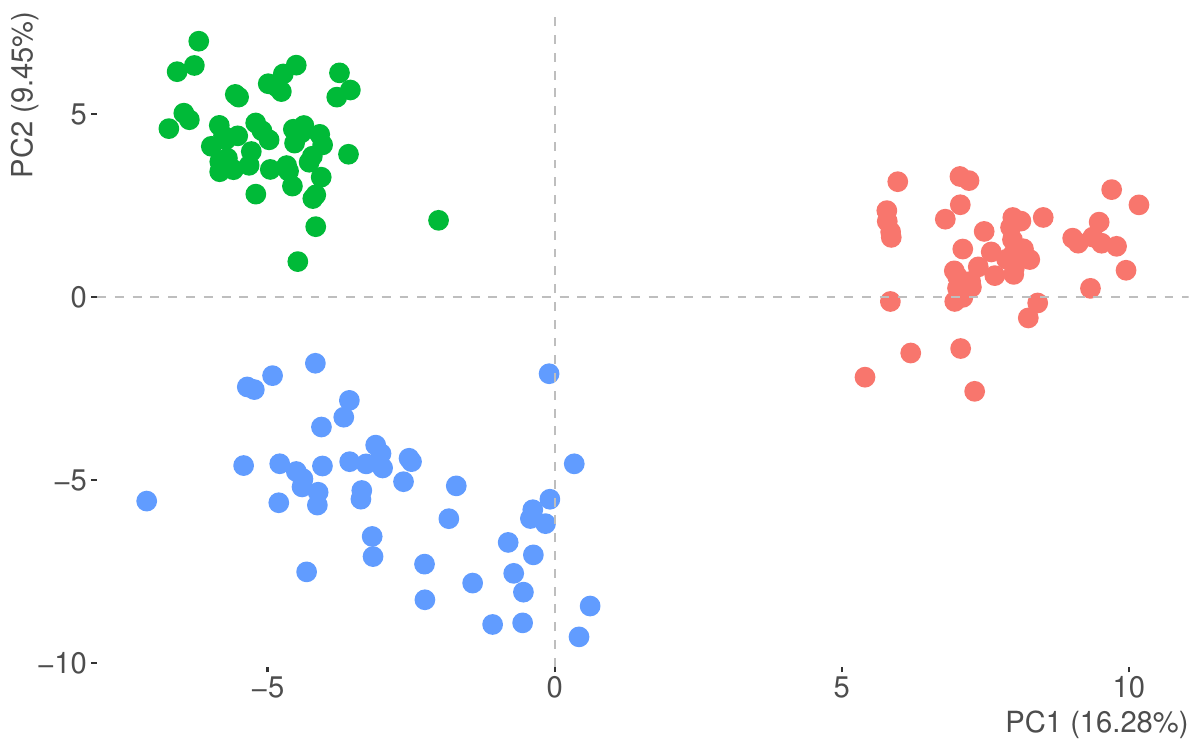}
\caption{\centering PCA of samples from Aquinas (red), Courson (green), and~Langton~(blue).}
\label{fig_2}
\end{figure}

Two factors limit the usefulness of the above test for the analysis of stylistic clusters within Langton's collection. First, since we have no reliable estimate of the number of expected classes, PCA alone is not a suitable clustering mechanism, as it can conflate some clusters discernible in the initial data. The second limitation is related to the samples' length. For the distributions of the most frequent words to be representative of the authorial signal, the sample length needs to reach a threshold of 2,000 to 5,000 words, with the exact required length varying depending on genre and language \citep{Eder2013}. Meanwhile, the average length of a single \textit{quaestio} in Langton's collection is around 1400 words, with the extreme values of 166 and 7385 words.\footnote{The numbers reported here and in Table 1 are representative of all published or preliminarily edited \textit{quaestiones}, which cover roughly 90\% of the entire material. The ongoing critical edition of \textit{quaestiones} (Langton, ed. Bieniak et al., 2014--2024) is planned for six volumes, four of which are already published, and the remaining two are at an advanced stage.} We can reach the reliable sample's length by concatenating the \textit{quaestiones} --- as in the above test --- but this effectively averages over the stylistic signal of all \textit{quaestiones} included in a given sample, obscuring the signals of shorter texts and under-representing the actual stylistic variance of the collection. 

While the most promising way to address this issue seems to be by extending the set of analysed features --- see the discussion in the 'Methods' section below --- this problem can be to some extent mitigated by bundling the \textit{quaestiones} according to the information obtained from stemmatical analysis. As already noted, the \textit{quaestiones} are transmitted in four subcollections, which contain different, partially overlapping sets of \textit{quaestiones}. Since these subcollections most likely originated as compilations of dispersed Langtonian material, it makes sense to analyse smaller classes of \textit{quaestiones} organized by the set of manuscripts in which they are transmitted. In this way, we end up with 10 disjoint classes, as detailed in Table 1. This organization of material accounts for major stemmatical relations, including the shifting relation between ms. C and family~$\gamma$.\footnote{As noted in the stemma, Ca-Cb, unlike Cc-Cf, are independent from $\gamma$. This organization of the material could be further improved by accounting for differences between H\,/\,K and $\alpha$, as well as distinguishing between sections of ms. C transmitting material found also in $\beta$. Unfortunately, some of such classes would score below 3000 words.} 

\begin{table}[hbt]
\begin{threeparttable}
\caption{Grouping Langton's \textit{Quaestiones} by shared codices}
\label{table_example}
\begin{tabular}{llll}
\toprule
\headrow Class by transmitting mss. & N of \textit{quaestiones}  & total length (in words)\\
\midrule
$\beta$ & 98 & 106221 \\
\midrule
$\gamma$ with Cc -- Cf & 65 & 86543 \\
\midrule
$\beta$ + $\gamma$ + C (any section) & 32 & 76215 \\
\midrule
Cb & 54 & 70282 \\
\midrule
$\gamma$ + Cb & 27 & 36776 \\
\midrule
$\gamma$ without C & 23 & 27113 \\
\midrule
$\beta$ + C (any section) & 12 & 23769 \\
\midrule
Ca & 9 & 18313 \\
\midrule
$\gamma$ + Ca & 9 & 17251 \\
\midrule
H\,/\,K & 11 & 11125 \\
\bottomrule
\end{tabular}
\begin{tablenotes}[hang]
\item[] For more details on this data, consult the supplementary files --- see the Data Availability Statement below.
\end{tablenotes}
\end{threeparttable}
\end{table}

\enlargethispage{2\baselineskip}
Figure 3 shows the results of PCA conducted for these classes, based on the distribution of the 200 most frequent words. While most classes expectedly cluster around the average for the entire collection, there are two clear outliers: the material transmitted exclusively in section Ca of ms. C, as well as \textit{quaestiones} proper to the Chartres collection H\,/\,K.\footnote{Notably, the classes displaying distinct stylistic signals are the shortest ones. The longer classes are also likely to contain portions of stylistically diverse material, but their location requires finer data granularity --- ideally at the level of individual \textit{quaestiones}.} In the case of Ca, this notably aligns with a long-standing palaeographic observation: the final folios of Ca --- the ones transmitting material not found in any other codices --- were copied by a different hand \citep{gregory1930cambridge}. Similarly, the bulk of \textit{quaestiones} transmitted solely by H\,/\,K is positioned on its final folios (ms. K, f. 152ra--153va), possibly also copied \textit{alia manu}.\footnote{Codex H\,/\,K was destroyed during the Second World War and is known today only through low-quality microfilm reproductions, rendering its palaeographic analysis at best tentative.} Thus, the exploratory analysis shows that even based on this admittedly unrefined set of features, it is possible to distinguish stylistic signal characteristic of this collection, as well as locate some stylistic heterogeneity within its boundaries.

\begin{figure}[hbt!]
\centering
\includegraphics[width=1\linewidth]{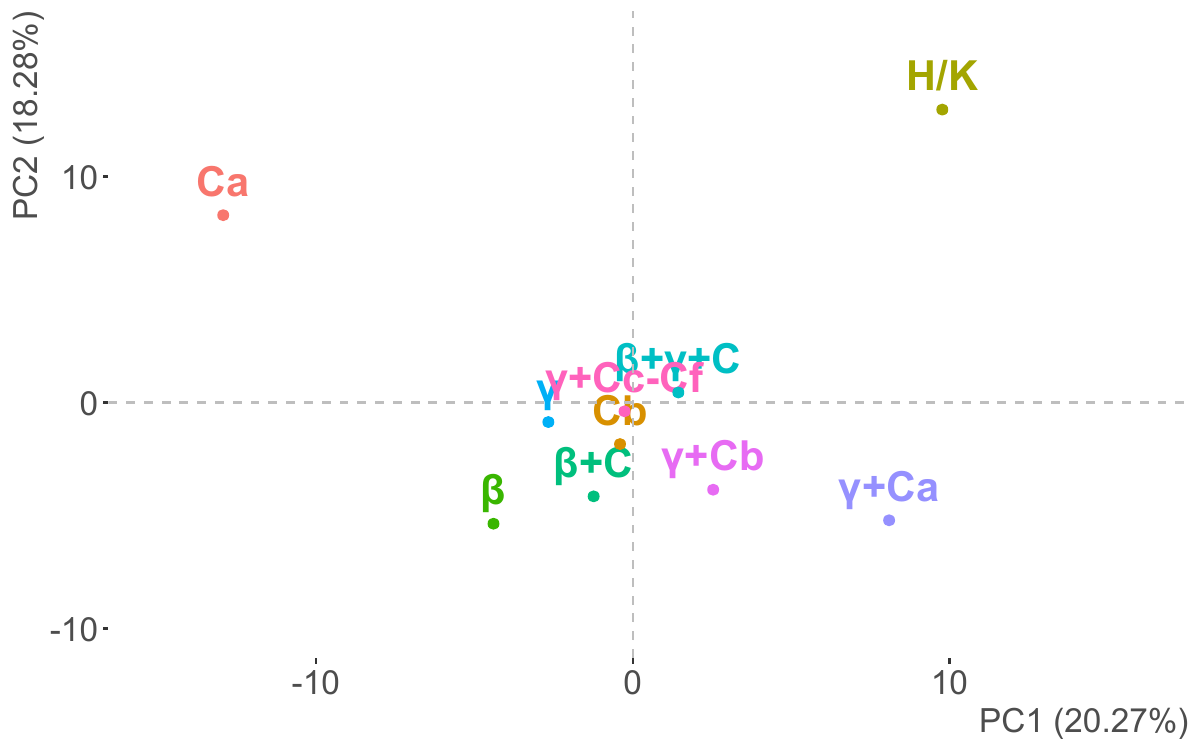}
\caption{PCA for Langton's Quaestiones, grouped by transmitting codices.}
\label{fig_3}
\end{figure}

\section{Methods}\label{methods}

The proposed study relies heavily on the methods applied in the context of similar research questions in \citet{CampsCléricePinche2021}, where a corpus of short and mostly anonymous Old French texts was analysed to uncover original authorial series obscured by layers of compilatory work. Moreover, this study demonstrated the validity of HTR-based data extraction pipelines for stylometric analysis. Below, I discuss the key implementation details of relevant stylometric tests and data preparation.

\subsection{Extended features: POS n-grams and pseudo-affixes}

While some stylometric tests proposed in recent literature perform well in authorship attribution tasks for samples much shorter than 3000 words, these solutions largely rely on word embeddings and training author-specific classifiers.\footnote{For examples, see the discussion of Multi-Author Writing Style Analysis Task at PAN 2024 --- \citet{CLEF2024}.} These techniques, in turn, require framing the problem as a supervised scenario based on a dataset of securely labelled samples, which is not feasible in this case. Instead, I plan to extend the set of analysed features, aiming to obtain richer representations of samples and thus enhance the performance on shorter \textit{quaestiones}.

A strategy suggested in some recent literature is to incorporate Part-of-Speech (POS) n-grams \citep{Chenetal}. Of many possible extended features, the POS 3-grams are especially promising as a simple representation of syntactic structures mostly ignored in the bag-of-words approach of tests based solely on word distributions. Incorporating POS 3-grams is further facilitated by the availability of efficient morphological taggers for Latin. For the proposed study, I intend to use the LatinPipe \citep{straka-etal-2024-ufal} or closely related UDPipe 2, both of which provide API access and report very high performance on POS tagging (over 99\% accuracy), including on scholastic Latin corpora.\footnote{For the reported performance, see \cite{straka-etal-2024-ufal}, Table 4, especially the performance on Index Thomisticus Treebank. The final paper will report performance measured on an annotated sample from Langton's collection.} Moreover, following \citet{CampsCléricePinche2021}, I will extend analysed features by pseudo-affixes, i.e. character 3-grams representing each word's boundaries,\footnote{To give an example, word 'verbum' would generate pseudo-affixes ‘\_ve’, ‘\^{}ver’, ‘bum\$’, and ‘um\_’.} which have been shown to provide valuable stylistic signals \citep{sapkota-etal-2015-character}.

For each individual feature (most frequent words, POS 3-grams, prefixes), the minimal statistically reliable sample length will be assessed implementing the test proposed by \citet{Moisl2011}, in which once more I follow \citet{CampsCléricePinche2021}. Establishing this threshold is the study's primary goal and will condition the later analysis of the data since it determines the exact set of \textit{quaestiones} which can be reliably subjected to cluster analysis.

\subsection{Data preparation}

For the proposed study, I will benefit from access to the machine-readable text of most or all of Langton's \textit{quaestiones}. Nevertheless, I also intend to perform tests on HTR-extracted transcriptions. It can be expected that no significant difference in performance will be observed, with the critical edition being effectively a denoising procedure, although we cannot \textit{a priori} rule out the possibility that editorial interventions left some systematic stylistic trace. The primary goal in experimenting with automated transcriptions is to develop workflows facilitating the inclusion of relevant unedited sources (or manuscript-specific versions of edited material) in further stylometric studies. In the immediate context of Langton's corpus, this would offer great aid in the survey of his vast and mostly unstudied scriptural commentaries. 

I intend to test in this study the relatively recent \linebreak transformer-based HTR solutions (TrOCR), which have been successfully applied to historical material \citep{strobel2022transformer}. These architectures rely on a vision transformer for feature extraction and a BERT-type decoder for the translation of visual tokens into characters, offering a few relevant advantages over widely applied solutions based on convolutional neural networks. First, they work exceptionally well with normalized transcriptions, largely facilitating the preparation of ground truth. This comes with a significant advantage in the context of university-based Latin literary production, which features a high density of often idiosyncratic abbreviations. In this case, framing the abbreviation expansion as a downstream task performed on HTR-extracted (semi-)diplomatic transcription is considerably more complex than for vernacular corpora, which generally confer less frequent and more regular abbreviations.\footnote{For a relevant example of transcription guidelines framing abbreviation expansion as a downstream task, see \citet{catmus}.} Moreover, the reliance on a transformer decoder is likely to result in noise reduction: since the model has a high preference for regular forms, it will likely at least partially normalize orthography, facilitating the later lemmatization task. Even where the transcription is inaccurate, the produced form can be sufficiently close to ground truth to enable correct assignment of POS tags and prefixes. Consequently, the task-specific accuracy of extracted features is likely to be significantly higher than suggested by the reported Character Error Rate of the model, which can be expected to score around 2--3\%.

For ground truth preparation, I plan to rely on Kraken’s blla model for text segmentation.\footnote{Documentation available at \url{https://kraken.re/main/api_docs.html}} While \citet{CampsCléricePinche2021} reported low performance for segmentation with Kraken’s legacy model (default at the time), initial tests show that currently blla outperforms Transkribus’ Universal Lines in polygonization, creating overall more spacious line polygons and capturing relevant abbreviation markers. I will reuse the transcriptions provided by the collection's editors, manually aligning a portion of the material (c.~20 pages), after which I will train a provisional Kraken model and automatically align the transcription for remaining pages.\footnote{Automatic transcription alignment, based on PASSIM script for text reuse detection, was implemented in eScriptorium 0.13. On these projects, see \citet{passim}, \citet{eScriptorium}.} While it would be convenient to prepare in this way ground truth for all major codices transmitting Langton's collection, I will prioritize workflow exploration over providing a comprehensive dataset. 

\section{Potential Results}\label{results}
As noted above, the final results of this study will depend heavily on the exact value of the minimal sample length established in statistical tests. It should be noted that this threshold is calculated for every individual feature and depends on the feature's overall probability in the corpus. Consequently, it will be necessary to balance out the exact set of features and corpus composition, almost certainly resulting in the exclusion of some of the shortest \textit{quaestiones}. Depending on the composition of the final corpus, the study will address three questions: 
\begin{itemize}
    
    \item[--] Can we discern some distinct clusters among longer \textit{quaestiones}? Such clusters would likely correspond to the activity of different editors, potentially including a cluster of \textit{quaestiones} directly corrected by Langton.
    
    \item[--] In general, do short and long versions of one \textit{quaestio} tend to cluster together? Such clusters could indicate cases in which either the original \textit{reportator} developed the longer version or in which the longer version preserved \textit{verbatim} most of the \textit{reportatio}. If no clusters of this type are observed, this would suggest a systematic stylistic difference between \textit{reportationes} and literary \textit{quaestiones} beyond the obvious difference in length.
    
    \item[--] Finally, if it will be possible to include most of the short \textit{quaestiones}, can we observe any clusters of \textit{reportationes}? Such clusters could be linked to the activity of individual \textit{reportatores}.
    
\end{itemize}









\newpage

\paragraph{Acknowledgments}

I am thankful to Magdalena Bieniak and Wojciech Wciórka for reading an earlier version of this paper and sharing their helpful remarks. I would also like to thank Gary Macy for sharing his transcription of ms. Bruges 247.

\paragraph{Funding Statement}
This work was supported by the National Science Centre, Poland, project 2022/45/N/HS1/03747.

\paragraph{Competing Interests}
The author declares none.

\paragraph{Data Availability Statement}

The data and code used in this study are available at: \href{https://github.com/jtmaliszewski/CHR-2025-It-takes-a-village}{https://github.com/jtmaliszewski/CHR-2025-It-takes-a-village}. Please note that due to unresolved copyright concerns, the plain text data was masked: all but the top 200 most frequent words were replaced with a 'MASKEDTOKEN' placeholder. This allows for full reproduction of the exploratory analysis presented in this paper, and the unmasked data was disclosed for peer review. I am currently seeking permission from relevant parties to publish the entire unmasked corpus as part of the final research report. In the meantime, if you are interested in inspecting the unmasked corpus, please contact me at \href{mailto: j.maliszewski@uw.edu.pl}{j.maliszewski@uw.edu.pl}.

The stylometric analysis employed in this paper was implemented with the \textit{stylo} package for R --- \citet{Eder2016-bc}. 

\paragraph{Ethical Standards}
The research meets all ethical guidelines, including adherence to the legal requirements of the study country.



\defbibnote{preamble}{By default, this template uses \texttt{biblatex} and adopts the Chicago referencing style. However, the journal you’re submitting to may require a different reference style; specify the journal you're using with the class' \texttt{journal} option --- see lines 1--9 of \emph{sample.tex} for a list of options and instructions for selecting the journal.}

\newpage

\section*{Primary sources}
\bigskip

\subsection*{Manuscripts}
\smallskip

\begin{small}
\smallskip\noindent
Stephen Langton, \textit{Quaestiones theologiae}

{\leftskip=10pt

 \noindent A \tabto{15pt} Avranches, Bibliothèque municipale, 230, ff.~12\textsuperscript{ra}--294\textsuperscript{rb}

 \noindent B \tabto{15pt} Arras, Bibliothèque municipale, 965 (394), ff.~70\textsuperscript{ra}--157\textsuperscript{vb}

 \noindent C \tabto{15pt} Cambridge, St. John's College Library, C.7 (57),\\ \tabto{15pt} ff.~171\textsuperscript{ra}--352\textsuperscript{rb}

 \noindent {\leftskip=25pt(Ca = C, ff.~171--218; Cb = C, ff.~219--282; Cc = C, ff.~283--306; \phantom{(}Cd = C, ff.~307--322; Ce = C, ff.~323--346; Cf = C, ff.~347--352)
 
 }

 \noindent H \tabto{15pt} Chartres, Bibliothèque municipale, 430, ff.~3\textsuperscript{r}--73\textsuperscript{v}

 \noindent K \tabto{15pt} Chartres, Bibliothèque municipale, 430, ff.~74\textsuperscript{ra}--154\textsuperscript{vb}

 \noindent L \tabto{15pt} Oxford, Bodleian Library, Lyell~42

 \noindent R \tabto{15pt} Città del Vaticano, Biblioteca Apostolica Vaticana,\\ \tabto{15pt} Vat. lat. 4297

 \noindent S \tabto{15pt} Paris, Bibliothèque nationale de France, lat. 16385

 \noindent V \tabto{15pt} Paris, Bibliothèque nationale de France, lat. 14556

}

\medskip \noindent
Robert of Courson, \textit{Summa}

{\leftskip=10pt

\noindent Bruges 247 = Brugge, Hoofdbibliotheek Biekorf (Stadsbibliotheek), 247

}

\end{small}

\medskip
\subsection*{Editions}
\medskip

\begin{small}
    \noindent \hangindent15pt Hugh of St. Victor, \textit{Sententiae de divinitate}. In Ambrogio Piazzoni, "Ugo di San Vittore auctor delle \textit{Sentetiae de divinitate}", Studi Medievali 23 (1982), 912--55. \medskip

   \noindent \hangindent15pt Peter Abelard, \textit{Historia Calamitatum}, ed. Jacques Monfrin, Paris: Vrin, 1959. \medskip

    \noindent \hangindent15pt Peter Comestor, Lectures on the Glossed Gospel of John, ed. and tr. David M. Foley, 2024. \medskip

\noindent \hangindent15pt Stephen Langton, \textit{Quaestiones theologiae}, Auctores Britannici Medii Aevi (ABMA):

\leftskip = 10pt    

\begin{itemize}[label={}]
    \item Vol. I, ed. Riccardo Quinto, Magdalena Bieniak, 2014, \\ \hspace*{\fill} (ABMA~22)

    \item Vol. II, ed. Wojciech Wciórka, \textit{in preparation}\\ \hspace*{\fill} \phantom{(ABMA)}

    \item Vol. III.1, ed. Magdalena Bieniak, Wojciech Wciórka, 2021, \\ \hspace*{\fill} (ABMA~36)

    \item Vol. III.2, ed. Magdalena Bieniak, Marcin Trepczyński, Wojciech \\ \hspace*{\fill} Wciórka, 2022, (ABMA~40)

    \item Vol. III.3, ed. Magdalena Bieniak, Andrea Nannini, 2024,\\ \hspace*{\fill} (ABMA~45)

    \item Vol. IV, ed. Magdalena Bieniak, Jan Maliszewski, \textit{in preparation}
\end{itemize}
\end{small}

\begin{small}

\noindent \hangindent15pt Thomas Aquinas, \textit{Summa Theologiae}, \textit{prima pars} (= \textit{Opera omnia iussu impensaque Leonis XIII P. M. edita}, t. 4-5, Roma 1888--1889). Digitised text by R. Busa, E. Alarcón is available from \href{https://www.corpusthomisticum.org/}{Corpus Thomisticum}.

\end{small}

\medskip

\printbibliography[title = {Other references}]




\end{document}